\documentclass[conference]{IEEEtran}
\IEEEoverridecommandlockouts

\usepackage[letterpaper,left=0.64in,right=0.64in,top=0.75in,bottom=1.05in]{geometry}

\usepackage{cite}
\usepackage{amsmath,amssymb,amsfonts}
\usepackage{algorithmic}
\usepackage{graphicx}
\usepackage{textcomp}
\usepackage{xcolor}
\usepackage{multirow}
\def\BibTeX{{\rm B\kern-.05em{\sc i\kern-.025em b}\kern-.08em
    T\kern-.1667em\lower.7ex\hbox{E}\kern-.125emX}}
\begin{document}

\title{A Hybrid Intelligent Framework for Uncertainty-Aware Condition Monitoring of Industrial Systems %
\thanks{\IEEEauthorrefmark{1}Corresponding author}}


\author{\IEEEauthorblockN{Maryam Ahang}
\IEEEauthorblockA{
\textit{University of Victoria}\\
Victoria, Canada \\
maryamahang@uvic.ca}
\and
\IEEEauthorblockN{Todd Charter}

\IEEEauthorblockA{
\textit{University of Victoria}\\
Victoria, Canada \\
toddch@uvic.ca}
\and
\IEEEauthorblockN{Masoud Jalayer}
\IEEEauthorblockA{
\textit{Aalto University}\\
Espoo, Finland \\
masoud.jalayer@aalto.fi}
\and
\IEEEauthorblockN{Homayoun Najjaran\IEEEauthorrefmark{1}}
\IEEEauthorblockA{
\textit{University of Victoria}\\
Victoria, Canada \\
najjaran@uvic.ca}

}

\maketitle

\begin{abstract}
Hybrid approaches that combine data-driven learning with physics-based insight have shown promise for improving the reliability of industrial condition monitoring. This work develops a hybrid condition monitoring framework that integrates primary sensor measurements, lagged temporal features, and physics-informed residuals derived from nominal surrogate models. Two hybrid integration strategies are examined. The first is a feature-level fusion approach that augments the input space with residual and temporal information. The second is a model-level ensemble approach in which machine learning classifiers trained on different feature types are combined at the decision level. Both hybrid approaches of the condition monitoring framework are evaluated on a continuous stirred-tank reactor (CSTR) benchmark using several machine learning models and ensemble configurations. Both feature-level and model-level hybridization improve diagnostic accuracy relative to single-source baselines, with the best model-level ensemble achieving a 2.9\% improvement over the best baseline ensemble. To assess predictive reliability, conformal prediction is applied to quantify coverage, prediction-set size, and abstention behavior. The results show that hybrid integration enhances uncertainty management, producing smaller and well-calibrated prediction sets at matched coverage levels. These findings demonstrate that lightweight physics-informed residuals, temporal augmentation, and ensemble learning can be combined effectively to improve both accuracy and decision reliability in nonlinear industrial systems.
\end{abstract}

\begin{IEEEkeywords}
Condition monitoring, Hybrid fault detection, Decision fusion, Machine learning, Uncertainty quantification.
\end{IEEEkeywords}

\section{Introduction}

Prognostics and Health Management (PHM) is essential for maintaining the safety, efficiency, and reliability of industrial plants. Common approaches include model-based fault detection, signal-based or data-driven fault detection, and hardware redundancy. Data-driven approaches are widely used due to their flexibility and strong empirical performance \cite{surucu2023condition}, but they rely heavily on the availability and quality of data and often fail to incorporate prior knowledge about process physics. Model-based methods, in contrast, use mathematical representations of the system to detect deviations from nominal behaviour \cite{isermann2005model}, although constructing accurate models for complex industrial processes can be difficult. Hybrid approaches that combine these paradigms have therefore gained increasing attention, since they can leverage complementary strengths to improve prediction accuracy and robustness \cite{wilhelm2021overview, ahang2024intelligent}. Such hybridization may occur at the data level, the algorithmic level, or the decision level \cite{von2021informed}.

Several studies have explored how model-based and data-driven information can be fused for improved diagnostic reliability. Tidriri et al. \cite{tidriri2018generic} proposed a decision fusion framework based on a discrete Bayesian network that integrates model-based and data-driven classifiers, demonstrating improved performance on the Tennessee Eastman process. Chao et al. \cite{chao2022fusing} developed a hybrid prognostics framework that uses physics-based models to infer unobservable parameters which are then combined with sensor data and processed by a deep neural network to estimate remaining useful life. Jung and Krysander \cite{jung2024assumption} studied the integration of model-based residuals with data-driven classifiers to support fault decoupling when faulty training data are scarce, highlighting the importance of combining design principles with learning-based methods. 

Ensemble learning provides another opportunity to improve reliability, since combining multiple classifiers can reduce variance and enhance predictive performance. Ensemble approaches such as voting, bagging, boosting, and stacking have demonstrated strong performance across many condition monitoring applications \cite{sagi2018ensemble, mian2024literature}. However, most data-driven ensemble methods do not express uncertainty in their predictions. In safety-critical domains, a lack of calibrated uncertainty can limit the practical applicability of otherwise accurate classifiers. Conformal prediction offers a distribution-free means of generating prediction sets with guaranteed coverage levels for uncertainty quantification \cite{shafer2008tutorial} and has recently been studied to increase decision reliability in fault diagnosis \cite{nemani2023uncertainty, heddoub2025uncertainty, wu2025query}.

Although prior work has shown the benefits of hybrid modelling and ensemble learning, the relative advantages of different hybrid integration strategies and their influence on uncertainty quantification are not yet well characterized. This is especially relevant in nonlinear closed-loop systems, where faults may be subtle, partially compensated by control action, or evolve gradually over time. Motivated by these challenges, this work investigates how physics-informed residuals, temporal feature augmentation, and ensemble learning can be integrated in a lightweight and practical manner to improve both diagnostic performance and uncertainty characterization.

In this paper, we develop and evaluate a hybrid monitoring framework with the following contributions:

\begin{itemize}
    \item A physics-informed residual generation method based on nominal surrogate models trained on a single normal operating run. These models remain fixed for all other conditions and provide a simple way to incorporate physical knowledge into the monitoring process.

    \item An investigation of two hybrid integration strategies that combine primary measurements, lagged temporal features, and physics-informed residuals. The first strategy operates at the feature level and the second at the model level through ensemble fusion.

    \item A comprehensive evaluation of these integration strategies on the CSTR benchmark, including the use of multiple machine learning models and ensemble configurations, to assess how different information sources contribute to overall diagnostic performance.

    \item An uncertainty analysis using conformal prediction that examines coverage, prediction-set characteristics, and abstention behavior, providing insight into how each hybrid strategy affects the reliability of fault decisions.
\end{itemize}

\section{Methodology}

This section describes the proposed hybrid condition monitoring framework, which combines data-driven learning, physics-informed residual modeling, ensemble inference, and uncertainty quantification. The methodology is designed to address challenging fault scenarios in nonlinear, closed-loop industrial processes, where faults may be subtle, partially compensated by control action, or evolve over time.

The proposed framework consists of four main components:
\begin{enumerate}
    \item data-driven fault classification using multiple lightweight supervised learning models,
    \item physics-informed residual feature extraction based on a nominal surrogate representation of the system,
    \item hybrid and ensemble fusion of data-driven and residual-based information to improve diagnostic robustness, and
    \item uncertainty quantification using conformal prediction to assess the confidence of model outputs.
\end{enumerate}

\begin{figure*}[t]
  \centering
  \vspace{0.08in}
  \includegraphics[width=.85\linewidth]{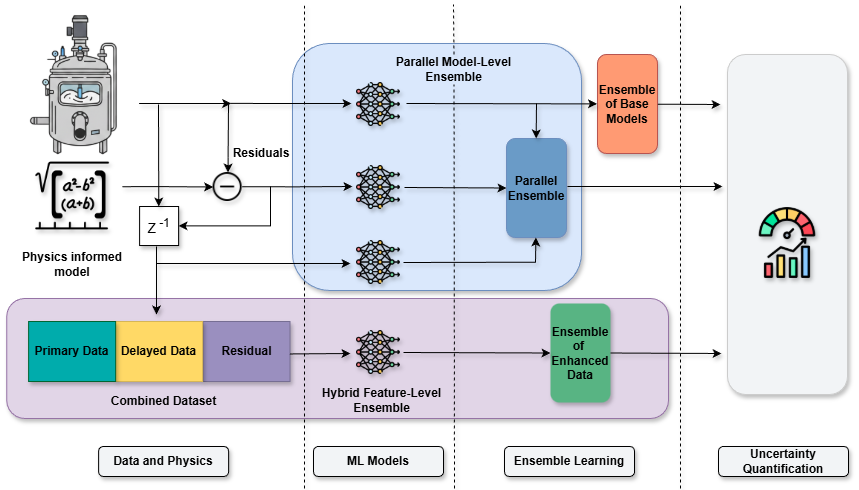}
  \caption{The framework of the proposed method (icons from Flaticon)}
  \label{Framework}
\end{figure*}

Figure \ref{Framework} provides an overview of the proposed framework, illustrating the interaction between raw measurements, feature construction, learning models, ensemble inference, and uncertainty-aware prediction.

\subsection{Problem Setup}
We consider a fault detection problem for a nonlinear continuous stirred tank reactor (CSTR) operating under feedback control. The system is monitored through seven measured variables: inlet concentration, reactor concentration, coolant flow rate, coolant temperature, inlet temperature, reactor temperature, and coolant inlet temperature. All measurements are sampled at a uniform rate.

Each sample is assigned to one of several operating conditions that include normal operation, incipient process faults, sensor drifts, abrupt sensor shifts, and stochastic disturbances as mentioned in Table~\ref{tb:cstr_faults}. Several faults evolve gradually over time, while others introduce sudden or random deviations. Due to the presence of feedback control, many faults are partially compensated, resulting in subtle deviations from nominal steady-state behavior. This makes fault discrimination based on instantaneous measurements challenging and motivates the use of temporal and physics-informed feature representations.

\begin{table}[hb]
\begin{center}
\caption{Fault scenarios of the CSTR process}\label{tb:cstr_faults}
\begin{tabular}{p{1cm} p{2.5cm} p{1cm} p{2.5cm}}
\hline
\textbf{Fault ID} & \textbf{Description} & \textbf{Value of $\delta$} & \textbf{Interpretation} \\ \hline
1  & $a = a_0 \exp(-\delta t)$      & 0.004 & Catalyst decay \\
2  & $b = b_0 \exp(-\delta t)$      & 0.005 & Heat transfer fouling \\
3  & $C_i = C_{i,0} + \delta t$     & 0.005 & Sensor shift \\
4  & $T_i = T_{i,0} + \delta t$     & 0.1   & Sensor shift \\
5  & $T_{ci} = T_{ci,0} + \delta t$ & 0.1   & Sensor shift \\
6  & $C = C_0 + \delta t$           & 0.005 & Sensor shift \\
7  & $T = T_0 + \delta t$           & 0.1   & Sensor shift \\
8  & $T_c = T_{c,0} + \delta t$     & 0.1   & Sensor shift \\
9  & $Q_c = Q_{c,0} + \delta$       & $-0.2$ & Sensor shift \\
10 & $C_i \sim \mathcal{N}(0, 0.005)$ & --  & Concentration disturbance \\
11 & $T_i \sim \mathcal{N}(0, 5)$     & --  & Reactant temperature disturbance \\
12 & $T_{ci} \sim \mathcal{N}(0, 5)$  & --  & Coolant temperature disturbance \\ \hline
\end{tabular}
\\[2mm]
\small Subscript 0 refers to the nominal value, which is 1 for both $a$ and $b$.
\end{center}
\end{table}

\subsection{Feature Construction}

To represent the process behavior under different fault scenarios, we construct three categories of features from the available measurements: primary measurements, lagged temporal features, and physics-informed residuals. These representations capture complementary aspects of the system dynamics.

\subsubsection{Primary Measurements}
The primary feature vector consists of the seven raw sensor measurements at each time step,
\[
x_t = [C_i, C, Q_c, T_c, T_i, T, T_{ci}]^\top .
\]
This representation captures the instantaneous process state but does not explicitly encode temporal dependencies or physical consistency.

\subsubsection{Lagged Temporal Features}

To capture the closed-loop dynamics of the CSTR under feedback control, each time sample is augmented with lagged observations of the base measurement vector $\mathbf{x}_t$ defined above. For a selected set of time lags $\mathcal{L} = \{1, 2, 5, 10\}$, a lag-augmented feature vector is constructed as
\[
\tilde{\mathbf{x}}_t = \big[\mathbf{x}_t,\; \mathbf{x}_{t-\ell_1},\; \mathbf{x}_{t-\ell_2},\; \dots,\; \mathbf{x}_{t-\ell_{|\mathcal{L}|}}\big].
\]

To ensure all lagged terms are well-defined, the first $L_{\max} = \max(\mathcal{L})$ samples of each run are discarded. Lagged features are computed independently for each simulation run, preventing information leakage across runs and preserving temporal coherence. This representation exposes delayed process responses and controller-induced transients that are not apparent from instantaneous measurements alone.

\subsubsection{Physics-Informed Residual Features}

While lagged features capture temporal dependencies in a data-driven manner, they do not explicitly encode physical consistency with the underlying process dynamics. A natural approach is therefore to introduce residual features that quantify deviations between measured behavior and a nominal process model. However, in this benchmark, the ``Ideal'' simulation operates at steady state, making simple residuals formed by subtracting ideal trajectories from noisy measurements largely equivalent to shifted versions of the original signals. To address this limitation, we approximate the system dynamics using a nominal surrogate model trained on a single Normal operating run. The residual generation process consists of three steps: estimation of time derivatives from measured signals, prediction of these derivatives using a parametric surrogate model, and computation of residuals as the difference between measured and predicted dynamics.

Using the available measurements $\{C_i, C, Q_c, T_c, T_i, T, T_{ci}\}$, we define three lightweight regression models that predict the time derivatives of key process states: coolant temperature $T_c$, reactor temperature $T$, and reactor concentration $C$. To represent reaction effects, we introduce a nonlinear reaction proxy $\phi_t$, defined as
\[
\phi_t = C_t \exp\!\left(-\eta / T_t\right),
\]
where $T_t$ is expressed in Kelvin and $\eta$ is a constant.

The nominal surrogate models are linear in their parameters:
\begin{align}
\widehat{\dot{T}}_{c,t} &= \beta_1\, Q_{c,t}(T_{ci,t}-T_{c,t}) + \beta_2 (T_t - T_{c,t}) + \beta_3, \\
\widehat{\dot{T}}_{t} &= \alpha_1 (T_{i,t}-T_t) + \alpha_2 (T_t - T_{c,t}) + \alpha_3 \phi_t + \alpha_4, \\
\widehat{\dot{C}}_{t} &= k_1 (C_{i,t}-C_t) + k_2 \phi_t + k_3.
\end{align}
These models are not intended to reproduce the full first-principles CSTR equations. Instead, they provide a consistent nominal dynamic predictor whose violations are informative under faults and disturbances.


Residual computation requires time derivatives of the measured signals. To reduce amplification of measurement noise, derivatives $\dot{C}$, $\dot{T}$, and $\dot{T}_c$ are estimated using a smoothed numerical differentiation scheme based on a Savitzky--Golay filter. Since the simulation output is uniformly sampled, a fixed sampling interval of $\Delta t = 1$ s is used throughout.

Let $\mathbf{z}_t^{(c)}$, $\mathbf{z}_t^{(T)}$, and $\mathbf{z}_t^{(C)}$ denote the regressor vectors associated with the coolant, temperature, and concentration models, respectively. Using a single Normal baseline run, the model parameters are estimated via ordinary least squares:
\begin{align}
\beta &= \arg\min_\beta \sum_t \big(\dot{T}_{c,t} - \mathbf{z}_t^{(c)\top}\beta\big)^2, \\
\alpha &= \arg\min_\alpha \sum_t \big(\dot{T}_{t} - \mathbf{z}_t^{(T)\top}\alpha\big)^2, \\
k &= \arg\min_k \sum_t \big(\dot{C}_{t} - \mathbf{z}_t^{(C)\top}k\big)^2.
\end{align}
The resulting parameters are frozen and applied unchanged to all subsequent runs and fault scenarios.

For a given run, predicted derivatives are computed using the frozen nominal parameters, and residual signals are defined as
\begin{align}
r_{T_c,t} &= \dot{T}_{c,t} - \widehat{\dot{T}}_{c,t}, \\
r_{T,t} &= \dot{T}_{t} - \widehat{\dot{T}}_{t}, \\
r_{C,t} &= \dot{C}_{t} - \widehat{\dot{C}}_{t}.
\end{align}
These residuals quantify violations of nominal behavior induced by faults, disturbances, or parameter drift. To capture the temporal evolution of these violations, lagged versions of the residual signals are constructed using the same lag set $\mathcal{L}$ as for the raw measurements. Residual features are aligned with the lagged measurement features by discarding the initial samples affected by lag construction, ensuring consistent sample counts across all feature sets.

\subsection{Machine Learning Models}
Fault classification is performed using several supervised learning models, including Random Forest (RF), Extreme Gradient Boosting (XGBoost), CatBoost, and a shallow multilayer perceptron (MLP). These models are selected for their strong performance on industrial data, relatively low computational complexity, and robustness when trained on moderate-sized datasets.

RF, XGBoost, and CatBoost are tree-based ensemble methods that construct collections of decision trees using different bagging and boosting strategies. Random Forests rely on bootstrap aggregation to reduce variance, whereas XGBoost and CatBoost employ gradient boosting to iteratively improve weak learners. CatBoost further incorporates ordered boosting and regularization strategies that reduce prediction bias and improve stability. These properties make tree-based models well suited to fault detection tasks involving heterogeneous sensor measurements.

In addition to tree-based methods, a shallow three-layer MLP is employed to capture nonlinear relationships among process variables that may not be explicitly represented by tree-based decision boundaries. The network architecture is intentionally kept shallow to mitigate overfitting and to remain consistent with practical industrial constraints.

More complex deep learning architectures are not considered, as they typically require large labeled datasets to generalize effectively. Instead, the selected models provide a balance between representational capacity, interpretability, and generalization performance.

\subsection{Ensemble Approaches}

To aggregate predictions from multiple base learners, several ensemble methods are evaluated, including Soft Average Voting, Weighted Soft Voting, Hard Voting, and Stacking with a Logistic Regression or Random Forest meta-learner. Soft Voting algorithms combine class probability estimates, while Hard Voting relies on majority voting of predicted labels. Stacking-based methods employ a higher-level meta-learner to optimally combine base model outputs, enabling adaptive weighting and improved fault discrimination.

To effectively combine the strengths of model-based and data-driven approaches, this work investigates two ensemble strategies. Both strategies aim to enhance fault detection performance and robustness by using residual information alongside raw sensor measurements and lagged features.

By comparing these ensemble strategies, the proposed hybrid framework identifies robust fusion mechanisms that improve classification accuracy and reliability under limited and noisy industrial data conditions.

\subsubsection{Hybrid Feature-Level Ensemble}

The first ensemble method combines information from the primary dataset, lagged features, and residuals from the physics-informed approach at the feature level to produce an enhanced dataset.

This enriched dataset captures the absolute system behavior and deviations from the nominal model, along with delayed data from previous samples to better account for system dynamics. The augmented feature vectors are used as inputs to the data-driven classifiers. Each classifier independently learns fault-related patterns from the hybrid feature space. The final decision is obtained by ensembling the outputs of all base models, thereby leveraging diverse decision boundaries and improving generalization.

\subsubsection{Parallel Model-Level Ensemble}

In the second approach, a parallel ensemble structure is adopted to combine the information from the independently trained models. The residual signals are used directly as inputs to machine learning models, while separate models are trained on the primary sensor measurements and lagged data independently, making three sets of predictions. As a result, for each data sample, three parallel prediction streams are generated.

The outputs of these parallel models are subsequently fused at the decision level using ensemble techniques. This strategy enables independent learning from the model discrepancies and primary process data, thereby preserving the distinct diagnostic characteristics of each information source.

\subsection{Uncertainty Quantification}
In industrial fault monitoring applications, point predictions alone are often insufficient for reliable decision making. Operators require not only a predicted fault label, but also an assessment of the model’s confidence in that prediction. This is particularly important in the presence of incipient faults, sensor drifts, and partially compensated disturbances, where classification uncertainty can be high even when overall accuracy appears satisfactory.

To quantify predictive uncertainty in a statistically principled manner, we employ conformal prediction, which provides set-valued predictions with guaranteed coverage under minimal assumptions. Conformal prediction is model-agnostic, distribution-free, and naturally applicable to multiclass classification problems, making it well suited to the fault monitoring setting considered in this work.

\section{EXPERIMENT AND ANALYSIS}

This section describes the CSTR case study, experimental setup, and performance evaluation of the proposed hybrid monitoring framework. Random Forest, XGBoost, CatBoost, and a shallow multilayer perceptron (MLP) are used as baseline classifiers, and their outputs are combined using ensemble learning. Hybrid monitoring is achieved by computing residual features from a physics informed model and lagged features and integrating them either at the decision level (parallel model level ensemble) or at the feature level (hybrid feature level ensemble). Model performance is evaluated using standard classification metrics and conformal prediction to quantify uncertainty.

\subsection{CSTR Case Study Description}
The proposed models are evaluated on a Continuous Stirred Tank Reactor (CSTR) process, which is a common unit in many chemical plants. The benchmark used in this paper was designed in Simulink by Pilario and Cao \cite{pilario2018canonical}. The process can be described by the following equations:

\begin{align}
\frac{dC}{dt} &= \frac{Q}{V}\left(C_i - C\right) - a k C + v_1 \label{eq:C_balance} \\
\frac{dT}{dt} &= \frac{Q}{V}\left(T_i - T\right) - a \frac{\Delta H_r}{\rho C_p} k C 
                - b \frac{UA}{\rho C_p V}\left(T - T_c\right) + v_2 \label{eq:T_balance} \\
\frac{dT_c}{dt} &= \frac{Q_c}{V_c}\left(T_{ci} - T_c\right)
                  + b \frac{UA}{\rho_c C_{pc} V_c}\left(T - T_c\right) + v_3 \label{eq:Tc_balance}
\end{align}

Where $Q$ and $Q_c$ denote the inlet and outlet flow rates of the coolant, respectively, and $T_{ci}$ and $T_c$ are the corresponding temperatures.
The variables $T_i$ and $T$ represent the inlet and outlet temperatures of the tank, and $C_i$ and $C$ denote the inlet and outlet concentrations of reactant $A$. 
The parameters $V$ and $V_c$ correspond to the volumes of the reactor and the jacket, respectively. 
Moreover, $\rho$ and $\rho_c$ are the fluid densities in the tank and jacket, and $C_p$ and $C_{pc}$ are the associated heat capacities. 
The term $UA$ denotes the heat transfer coefficient, and $\Delta H_r$ represents the heat of reaction. 
The reaction rate constant $k$ follows the Arrhenius law, given by $k = k_0 \exp\!\left(-\frac{E}{RT}\right)$. 
The additive terms $v_i$ model Gaussian noise. Measurement noise was also added to the sensors.
The system model considers $[C_i,\, T_i,\, T_{ci}]$ as inputs and $[C,\, T,\, T_c,\, Q_c]$ as outputs, which correspond to the monitored process variables \cite{li2020transfer}.

The faults are described in Table \ref{tb:cstr_faults}. Ten simulation runs are performed for each operating condition (12 faults and normal operation), with 1000 samples per run, to generate a statistically diverse dataset. This results in a total of 130,000 samples. The dataset is balanced, as each fault type contributes an equal number of simulation runs. Seven runs were used for training, and one run each for validation, calibration (for conformal prediction), and testing.


\begin{table}[t]
\caption{Performance Comparison of Base and Ensemble Models on the Primary Data}
\label{tab:performance_primary}
\centering
\setlength{\tabcolsep}{2pt}
\renewcommand{\arraystretch}{1.15}
\begin{tabular}{llcccc}
\hline
\textbf{Models} & \textbf{Model} & \textbf{Accuracy} & \textbf{Precision} & \textbf{Recall} & \textbf{F1-score} \\
\hline
\multirow{4}{*}{ML Models }
 & RF & 0.9549 & 0.9597 & 0.9549 & 0.9562 \\
 & XGBoost   & 0.9587 & 0.9621 & 0.9587 & 0.9597 \\
 & CatBoost   & 0.9560  & 0.9608 & 0.9560 & 0.9574 \\
 & MLP       & 0.9507 & 0.9620 & 0.9507 & 0.9533 \\
\hline
\multirow{5}{*}{Ensemble }
 & Soft Voting          & 0.9596 & 0.9644 & 0.9596 & 0.9608 \\
 & Weighted Soft Voting & 0.9596 & 0.9644 & 0.9596 & 0.9608 \\
 & Hard Voting          & 0.9578 & 0.9641 & 0.9578 & 0.9595            \\
 & Stacking (LogR) & \textbf{0.9605} & \textbf{0.9652} & \textbf{0.9605} & \textbf{0.9615} \\
 & Stacking (RF)       & 0.9574 & 0.9623 & 0.9574 & 0.9581 \\
\hline
\end{tabular}
\end{table}

\subsection{Experimental Setup}
Four machine learning models are used as base classifiers: Random Forest (RF), XGBoost (XGB), CatBoost, and a shallow multilayer perceptron (MLP). These models are selected to cover both tree-based and neural network paradigms while maintaining moderate model complexity suitable for industrial monitoring applications.

The Random Forest model consists of 200 decision trees. XGBoost is configured with 100 estimators. CatBoost is implemented with 500 boosting iterations, a learning rate of 0.05, and a maximum tree depth of 6. The MLP consists of three fully connected hidden layers with 64, 32, and 8 neurons, respectively, and is trained using the Adam optimizer with a learning rate of $10^{-3}$. All hyperparameters were selected based on preliminary testing. The performance comparison of the models trained on the primary data is shown in \ref{tab:performance_primary}.

To assess predictive uncertainty, conformal prediction is applied to probabilistic model outputs. Conformal prediction sets are constructed for all soft-voting and stacking-based ensembles using a held-out calibration dataset. Empirical coverage, average prediction set size, and the number of empty prediction sets are reported for multiple miscoverage levels $\alpha \in \{0.01, 0.05, 0.1, 0.15, 0.2\}$.

For each sample $(x_i, y_i)$ in the held out calibration dataset, we calculate a non-conformity score $s_i$. Following the standard predictor approach for multi-class problems, the score is defined as:
    \begin{equation}
        s_i = 1 - \hat{P}(y_i | x_i)
    \end{equation}
where $\hat{P}(y_i | x_i)$ is the softmax probability assigned to the true class $y_i$ by the underlying ensemble model. This score quantifies the model's error or lack of confidence in the correct label.


All experiments are conducted using Python 3.11 on a workstation equipped with an NVIDIA GeForce RTX 3090 GPU and an AMD Ryzen 9 CPU with 64 GB of memory.

\subsection{Hybrid Fault Detection Approaches}
Two hybrid strategies are evaluated to integrate residual information with data-driven fault detection models.

\subsubsection{Hybrid Feature-Level Ensemble}

\begin{figure}[!t]
\vspace{0.08in}
\centerline{\includegraphics[width=\linewidth]{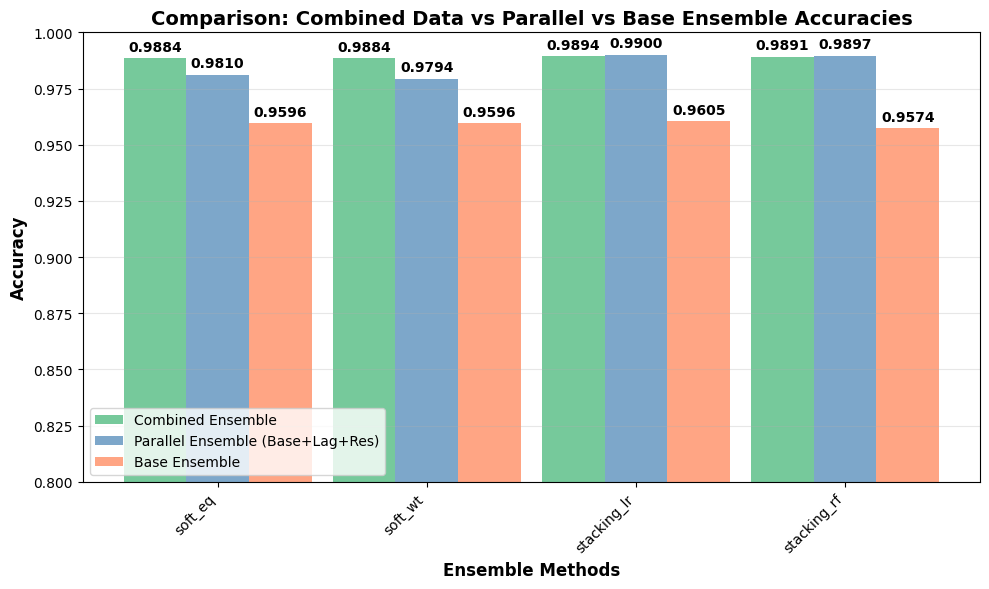}}
\caption{Accuracy comparison of different ensemble models}
\label{ensm}
\end{figure}

In this method, the integration of the information is done at the feature and data level. The residuals and lagged data are added to the primary dataset, and ML fault detection algorithms are trained on the augmented data. Using the augmented data, the accuracies of XGBoost, Random Forest, CatBoost, and MLP increase by about 3\%, 3\%, 3.1\%, and 2.4\%, respectively. Different ensemble learning methods are also applied to the models trained by the enhanced data, and there is an increase of 2.9\% compared to the primary on average. The summary of the fault detection metrics is presented as Feature-Level in the Table \ref{tab:feature_hybrid} and Table \ref{tab:performance_parallel}, and the comparison of different ensemble models is shown in Figure \ref{ensm}.

\begin{table}[t]
\centering
\caption{Fault classification performance using feature-level hybrid enhancement (Stacking (LogR)).
Residuals and lagged measurements are integrated directly into the feature space.}
\label{tab:feature_hybrid}
\begin{tabular}{l c c c c}
\hline
\textbf{Feature Set} & \textbf{Accuracy} & \textbf{Precision} & \textbf{Recall} & \textbf{F1-score}\\
\hline
Base                   & 0.9605 & 0.9652 & 0.9605 & 0.9615 \\
Base + Lag             & 0.9834 & 0.9848 & 0.9834 & 0.9837 \\
Base + Residual        & 0.9789 & 0.9806 & 0.9789 & 0.9793 \\
Base + Lag + Residual  & \textbf{0.9894} & \textbf{0.9900} & \textbf{0.9894} & \textbf{0.9896} \\
\hline
\end{tabular}
\end{table}

\subsubsection{Parallel Model-Level Ensemble}

To integrate information from the system's mathematical model with data-driven methods, the first approach was to perform fault detection on the primary data, the residuals, and lagged data independently.

The ML models introduced in the previous section were then trained independently on the primary data, their corresponding residuals, and the lagged data. In the next step, the models' predictions are combined using the introduced ensemble algorithms to produce the final decision. Compared to the ensemble of models trained on the primary data, the parallel hybrid approach increased accuracy by 2.9\% on average. The summary of the results is shown in Table \ref{tab:performance_parallel}.

\begin{table}[t]
\caption{Performance Comparison of Different Ensemble Models of Model-based and Data-driven methods.}

\label{tab:performance_parallel}
\centering
\setlength{\tabcolsep}{2pt}
\renewcommand{\arraystretch}{1.15}
\begin{tabular}{llccccc}
\hline
\textbf{Ensemble} & \textbf{Model} & \textbf{Accuracy} & \textbf{Precision} & \textbf{Recall} & \textbf{F1-score} \\
\hline
\multirow{5}{*}{Feature-Level}
 & Soft Voting          & 0.9884 &	0.9894	& 0.9884 &	0.9886 \\
 & Weighted Soft Voting & 0.9884 &	0.9894	& 0.9884 &	0.9886 \\
 & Hard Voting          & 0.9865 &	0.9880 & 0.9865 &	0.9869 \\
 & Stacking (LogR) & \textbf{0.9894} &	\textbf{0.9900} &	\textbf{0.9894} &	\textbf{0.9896} \\
 & Stacking (RF)       & 0.9891	& 0.9895 &	0.9891 &	0.9892 \\
 \hline
 \multirow{5}{*}{Parallel} & Soft Voting       & 0.9810           & 0.9834 & 0.9810           & 0.9816 \\
  & Weighted Soft Voting & 0.9794 & 0.9822 & 0.9794 & 0.9800 \\
 & Hard Voting          & 0.9746 & 0.9790 & 0.9746 & 0.9755 \\
 & Stacking (LogR) & \textbf{0.9900}           & \textbf{0.9906} & \textbf{0.9900} & \textbf{0.9901} \\
 & Stacking (RF) & 0.9897 & 0.9903     & 0.9897 & 0.9898 \\
\hline
\end{tabular}
\end{table}

\subsection{Performance Evaluation}

Models trained on base measurements already achieve strong accuracy, indicating that the monitored process variables are highly informative for fault detection. Incorporating lagged features and residuals leads to a substantial improvement in performance across all model types.

Models trained exclusively on residual features exhibit significantly lower classification accuracy. This outcome is expected, as residuals primarily capture model mismatch and abnormal behavior rather than explicit fault identity. Nevertheless, residual information plays an important role when integrated with other feature types.

Both the parallel model-level ensemble and the hybrid feature-level ensemble approaches improve fault detection performance compared to base models. The parallel model achieves the highest classification accuracy (99\%), while the feature level fusion also provides substantial gains over the primary datasets.

\subsection{Uncertainty and Conformal Prediction Analysis}
Conformal prediction is used to assess the framework's confidence. The models already show very high performance and accuracy, and the conformal prediction results demonstrate that empirical coverage closely matches the target confidence levels across all models and feature representations, confirming proper calibration and the absence of data leakage. However, substantial differences are observed in prediction set size and abstention behavior.

Residual-based models produce significantly larger average prediction sets and virtually no empty sets, indicating high epistemic uncertainty and ambiguity among fault classes. In contrast, models trained on base, lagged, and combined features tend to produce smaller prediction sets but exhibit higher rates of empty sets at moderate miscoverage levels, reflecting confident abstention when uncertainty exceeds acceptable risk thresholds.

In the parallel hybrid ensemble, uncertainty is influenced by the inclusion of residual-based models at the decision level. Since residual models naturally produce flatter probability distributions, their contribution leads to more conservative ensemble predictions, particularly under equal-weight soft voting. Weighted soft voting partially mitigates this effect by reducing the influence of less discriminative models.

\begin{figure}[!t]
    \centering
    \vspace{0.08in}
    \includegraphics[width=0.9\linewidth]{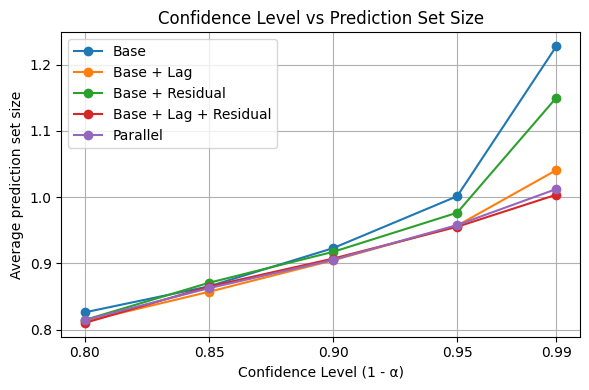}
    \caption{Confidence and prediction set size comparison using stacking (LogR) ensemble}
    \label{fig:cp}
\end{figure}

Feature-level hybrid enhancement yields the most efficient uncertainty behavior. By allowing residual information to modulate confidence within the classifier, combined-feature models achieve smaller conformal prediction sets for the same coverage level while maintaining appropriate abstention behavior. These results highlight the advantage of feature-level integration for uncertainty-aware fault monitoring. A comparison of the average prediction set size between different models is shown in Figure \ref{fig:cp}. This indicates that hybrid approaches reduce the average prediction set size to approximately 1, considering the very high classification accuracy around 99\%, suggesting that the model is precise and highly confident. Table \ref{tab:cp_performance} provides additional support for this conclusion, as it shows that adding the residuals ensures that all models meet the coverage requirement and have a small prediction set size. Although the differences are relatively small, the proposed parallel and feature-level ensemble methods with all feature types tend to give the best trade-off between coverage and prediction set size. It's worth noting that, since the conformal predictor may output empty sets (abstain), the average set size can be less than one. In our experiments, we noted that at higher confidence levels, very few empty sets were produced, with nearly zero at the 99\% confidence level.

\begin{table}[t]
\centering
\caption{Conformal prediction performance at different miscoverage levels.
Coverage (Cov.) and average prediction set size ($\boldsymbol{|C(x)|}$) quantify uncertainty behavior.}
\label{tab:cp_performance}
\renewcommand{\arraystretch}{1.15}
\setlength{\tabcolsep}{3.5pt}
\footnotesize
\begin{tabular}{l cc cc}
\hline
\textbf{Representation} 
& \multicolumn{2}{c}{$\boldsymbol{\alpha = 0.1}$} 
& \multicolumn{2}{c}{$\boldsymbol{\alpha = 0.01}$} \\
\cline{2-5}
& \textbf{Cov.} & $\boldsymbol{|C(x)|}$ 
& \textbf{Cov.} & $\boldsymbol{|C(x)|}$ \\
\hline
Base                    & 0.9103 & 0.9231 & 0.9919 & 1.2277 \\
Base + Lag              & 0.9034 & 0.9047 & 0.9896 & 1.0312 \\
Base + Residual         & 0.9149 & 0.9174 & 0.9939 & 1.1502 \\
Base + Lag + Res       & 0.9072 & 0.9073 & 0.9911 & 1.0037 \\
Parallel                & 0.9050 & 0.9054 & 0.9927 & 1.0127 \\
\hline
\end{tabular}
\end{table}

The experimental results demonstrate that residual features should not be interpreted solely in terms of classification accuracy. While residuals alone are insufficient for reliable fault detection, they provide valuable information about model consistency and abnormal system behavior. Conformal prediction reveals that residuals primarily affect uncertainty rather than discriminative performance.

From a practical perspective, the ability to abstain from uncertain decisions is preferable to confident misclassification in safety-critical systems. The proposed hybrid approaches, particularly feature-level integration, improve both fault detection accuracy and the reliability of uncertainty estimates, making them well suited for industrial condition monitoring applications.

\section{CONCLUSIONS}

This study investigated the impact of integrating data-driven and physics-informed methodologies on industrial condition monitoring. Using the CSTR benchmark, we demonstrated that a hybrid approach effectively bridges the gap between empirical learning and physical consistency.

To incorporate physical knowledge of the process, a physics-informed residual generation module was introduced, capturing deviations and their temporal derivatives relative to nominal operating conditions. Moreover, information from lagged measurements was incorporated to enrich the temporal context of the data. These complementary information sources were combined using two hybrid strategies: a feature-level ensemble approach and a parallel model-level ensemble approach, achieving 98.94\% and 99.00\% accuracy, respectively. These results correspond to an approximate 3\% improvement over standard ensemble methods, confirming that integrating heterogeneous information from multiple domains enhances fault detection and isolation capability beyond what can be achieved using process measurements alone.



Although classification metrics are important in condition monitoring, they are insufficient in understanding uncertainty and ensuring model confidence in decision-making. We further evaluated these approaches by quantifying uncertainty using conformal prediction, showing that the proposed models consistently satisfy coverage guarantees and produce prediction sets that are comparable in size to, or smaller than, those of the baseline models, making them practical for deployment in safety-critical industrial applications.
 
Future work will focus on testing the framework's scalability on more complex, large-scale industrial systems and real-world plants.

\section*{Acknowledgment} 
We would like to express our gratitude to FortisBC and MITACS for the financial support provided through the Grant Mitacs IT47118. 

\bibliographystyle{ieeetr}
\bibliography{bib}

\end{document}